\documentclass{article}
\usepackage[pdftex]{graphicx}

\graphicspath{{pdf/}{jpeg/}}
\DeclareGraphicsExtensions{.pdf,.jpeg,.png}
\usepackage{spconf,amsmath}
\usepackage{times}
\usepackage{epsfig}
\usepackage{amsmath}
\usepackage{amssymb}
\usepackage{tabularx}
\usepackage{cite}
\usepackage{makecell}
\usepackage{comment}
\usepackage{xcolor}

\title{In-bed Pressure-based Pose Estimation using Image Space\\Representation Learning}
\name{Vandad Davoodnia$^1$, Saeed Ghorbani$^2$, Ali Etemad$^3$
}
\address{$^{1,3}$Department of Electrical and Computer Engineering, Queen's University, Kingston, Canada\\
$^2$Department of Electrical Engineering \& Computer Science, York University, Toronto, Canada}
\begin{document}
\maketitle

\begin{abstract}
Recent advances in deep pose estimation models have proven to be effective in a wide range of applications such as health monitoring, sports, animations, and robotics. However, pose estimation models fail to generalize when facing images acquired from in-bed pressure sensing systems. In this paper, we address this challenge by presenting a novel end-to-end framework capable of accurately locating body parts from vague pressure data. Our method exploits the idea of equipping an off-the-shelf pose estimator with a deep trainable neural network, which pre-processes and prepares the pressure data for subsequent pose estimation. Our model transforms the ambiguous pressure maps to images containing shapes and structures similar to the common input domain of the pre-existing pose estimation methods. As a result, we show that our model is able to reconstruct unclear body parts, which in turn enables pose estimators to accurately and robustly estimate the pose. We train and test our method on a manually annotated public pressure map dataset using a combination of loss functions. Results confirm the effectiveness of our method by the high visual quality in the generated images and the high pose estimation rates achieved. 
\end{abstract}

\section{Introduction}
Sleeping makes up a third of human's life-span. As a result of recent advances in science, sleep studies, especially data-driven techniques, have attracted many researchers to the field. Moreover, low-cost processing and monitoring systems have enabled the utilization of sleep-related technologies in smart homes and clinics, paving the way for considerable impacts on health and quality of life.

Generally, sleep-related research includes the study of complications in the respiratory system, insomnia, and \textit{movement-related} disorders \cite{douglas2017guidelines}. Furthermore, it is shown that movement and posture during sleep have critical impacts on disorders such as sleep apnea \cite{lee2015changes} and pressure ulcers \cite{black2007national, liu2014bodypart}. As a result, monitoring posture in smart home and clinical settings is of great importance in order to identify or prevent the occurrence of such disorders.

Sleep monitoring technologies, such as textile-based pressure recording systems, have enabled ubiquitous and automated monitoring of movement, allowing for their use in both clinical health-care and research\cite{javaid2017balance, ostadabbas2014bed}. However, most of the studies using such systems are limited to coarse pose identification \cite{ostadabbas2014bed, zhao2020self, yousefi2011bed, davoodnia2020deep, davoodnia2019identity}, namely left, supine, and right postures. However, in clinics, it is critical to obtain information about specific pressure points and the relative pose of the limbs with respect to the body \cite{cunha2016neurokinect, peterson2010effects, walton2009prevention}. Consequently, in-bed body part localization and pose estimation has recently attracted researchers \cite{clever2020bodies}. Nonetheless, such related works are very scarce and very little work has been performed in this area.

\begin{figure}
\begin{center}
{\includegraphics[width=1\linewidth]{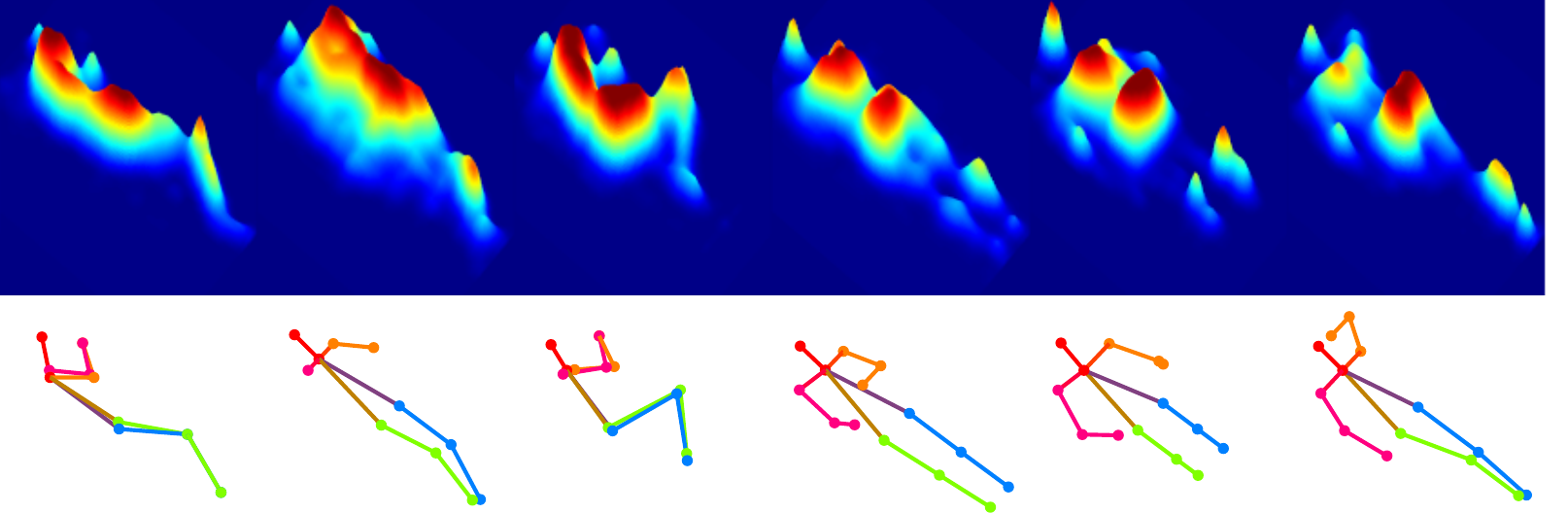}}
\end{center}
  \caption{
  Input pressure maps recorded using a mattress with embedded sensors is presented (top row) where weak and vague pressure points are observed.  The estimated poses using our proposed end-to-end method are presented (bottom row).}
\label{fig:teaser}
\end{figure}

With the recent advances in deep learning, a number of data-driven methods have been developed for pose estimation using natural camera-based images for a wide range of applications such as animation, robotics, sports, and human tracking \cite{tang2018deeply, ke2018multi, yang2017learning, cao2016realtime, chen2018cascaded}. Although these models are capable of achieving  strong body pose recognition, they perform poorly when used on matrices acquired through in-bed pressure-mapping systems, mostly due to the difference in postures and low-pressure points such as the head, knees, and hands. In this paper, we address this issue by proposing a learnable pre-processing block that enables off-the-shelf pose estimation models to generalize to pressure maps as well.

In this paper, we propose an end-to-end framework that allows off-the-shelf pose estimation models to be used for in-bed pose estimation, regardless of challenges such as weak, noisy, or vague pressure points (see Figure \ref{fig:teaser}). Our method includes two main modules, PolishNet, a fully-convolutional hourglass neural network that processes the input pressure maps such that the output lies within the subsequent conventional off-the-shelf pose estimator, pre-learned on camera-based images. We use OpenPose \cite{cao2016realtime} as the pre-learned pose estimator in our end-to-end pipeline.
In our architecture, we keep the parameters of the pose estimator frozen, while training PolishNet to minimize the Part Affinity Field (PAF) and heatmap losses between the annotated images and predicted outputs. We show that our end-to-end method out-performs the mere use of pose estimation methods significantly and detects limb positions robustly in a leave-one-subject-out validation strategy. Furthermore, we demonstrate that our network is capable of generating proper images for the pose estimators, even when trained on only a limited number of $10$ subjects, while single pose estimation models require large datasets to perform appropriately. Finally, to further evaluate the performance of our approach with other pose estimators, we swap OpenPose with another pose estimator, DeeperCut \cite{insafutdinov2016eccv, insafutdinov2017cvpr}, and observe robust performance, which points to the modular and generalized nature of our proposed solution.

\section{Proposed Method}
\noindent\textbf{Overview:} Our goal is to learn a pre-processing step that receives the pressure data as inputs and synthesizes images such that a pre-trained pose estimation module shows stable and accurate performance. In other words, the output data from the learner should lie on the data manifold used by the pose estimation module. Therefore, this learnable pre-processing step, which we call PolishNet, converts the pressure data to polished images that better resemble human figures as expected by the commonly available pose estimation models.

\begin{figure}
\begin{center}
{\includegraphics[width=\columnwidth]{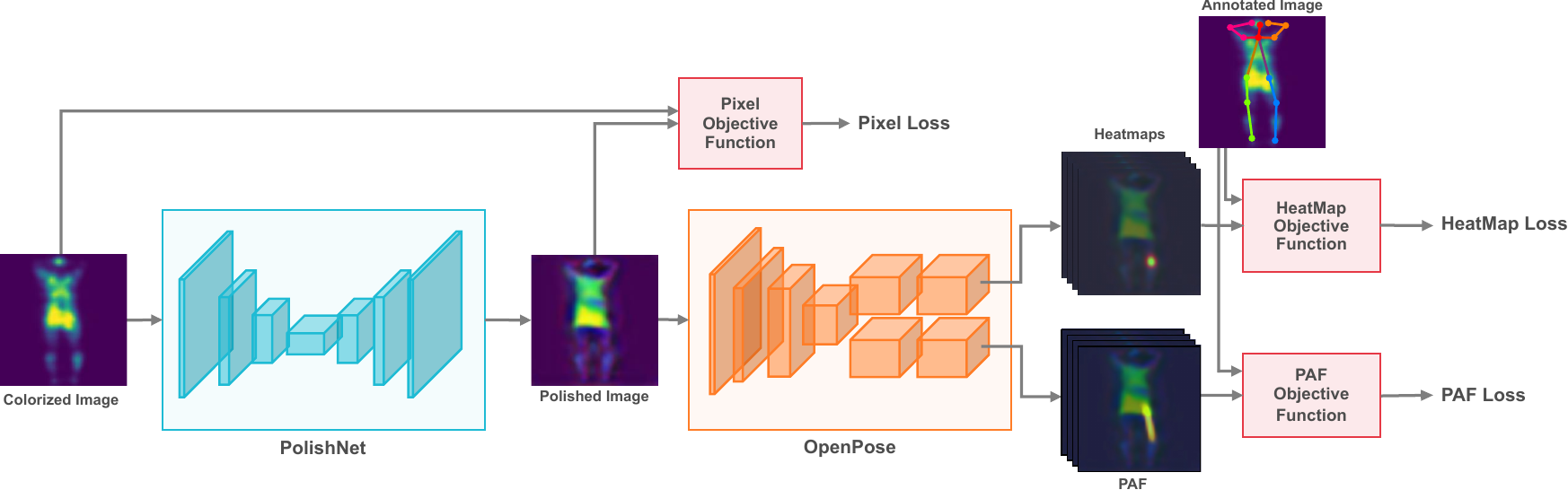}}
\end{center}
   \caption{Our proposed framework is presented. PolishNet is designed to learn the transformation from raw pressure maps into polished images, which can be fed directly to the pre-trained OpenPose module. The objective incorporates heatmap and PAF losses to force the PolishNet to synthesize completed body parts, as well as pixel loss to discourage the polished images to largely deviate from the pressure maps.}
\label{fig:network_architecture}
\end{figure}

Our pipeline consists of two blocks: (1) PolishNet which filters the pressure maps (2) A pre-trained pose estimation network that generates a heatmap for every body part that corresponds to their location. As illustrated in Figure \ref{fig:network_architecture}, PolishNet utilizes a combination of loss functions, namely the pixel mean-squared error of an image space between input and the PolishNet output, a heatmap loss corresponding to the body part positions, and a Part Affinity Fields (PAF) loss for body part identification. By training the pipeline using the mentioned loss functions, we ensure the generation of polished images consistent with the pose estimation networks' input data manifold while keeping the general properties of the input image.

\noindent\textbf{Architecture and Loss:} Let $I \in \mathbb{R}^{W \times H \times 3}$ be the input pressure data. Pre-processing the input is then performed by a variant of an hourglass network $\textit{P}$ called PolishNet. Our proposed network (see Figure \ref{fig:network_architecture}) contains three encoder blocks of Conv-Conv-BatchNorm-LeakyReLu and three blocks of DeConv-DeConv-BatchNorm-LeakyReLu on the decoder side. By utilizing the encoder-decoder blocks, we enable the network to capture the properties of the pressure data and incorporate pose and shape information in the latent space to generate the desired image in the polished data space $I'$, which is compatible with the pose estimation module. Accordingly, $I' = \textit{P}(I; \theta_{\textit{P}}), I' \in \mathbb{R}^{W \times H \times 3}$ is the output of PolishNet, where $\theta_{\textit{P}}$ are the network's trainable parameters.

To estimate the body pose and train PolishNet, we utilize OpenPose \cite{cao2016realtime}, a fast and reliable pose estimation network ($Q$) capable of accurately detecting different body joints known as keypoints. The output of OpenPose includes several heatmaps ($H$) and PAF ($F$) for each keypoint and its connections. Each heatmap is a $2$D distribution of the belief that a keypoint is located on each pixel, while the PAF is defined as a $2$D vector field connecting two limbs, encoding both position and the orientation of the connection. For our purposes, we only utilize the visible keypoints of the head, neck, shoulders, elbows, wrists, ankles, knees, and the hip, for a total of $h = 14$ heatmaps and $f = 28$ PAF for their connections. Accordingly, we define $H', F' = \textit{Q}(I'; \theta_{\textit{Q}})$, where $\theta_{\textit{Q}}$ is a set of trainable parameters, and $H' \in \mathbb{R}^{W_h \times H_h \times 14}$ and $H' \in \mathbb{R}^{W_h \times H_h \times 28}$ are the estimates of the ground-truth heatmaps and PAF respectively.

Next, we define an objective function with a heatmap term $E_{heatmap}$, a PAF term $E_{PAF}$, and a pixel loss term $E_{pixel}$ for training PolishNet as follows:
\begin{equation}
\begin{aligned}
E_{heatmap}   & = \sum_{k=1}^{h}{\sum_{i=1}^{H_h}{\sum_{j=1}^{W_h}{\left \| H_k - H_k' \right \|_{2}^{2}}}} \,\,\, ,
\end{aligned}
\end{equation}
\begin{equation}
\begin{aligned}
E_{PAF}       & = \sum_{k=1}^{f}{\sum_{i=1}^{H_h}{\sum_{j=1}^{W_h}{\left \| F_k - F_k' \right \|_{2}^{2}}}} \,\,\, ,
\end{aligned}
\end{equation}
\begin{equation}
\begin{aligned}
E_{pixel}     & = \sum_{i=1}^{H}{\sum_{j=1}^{W}{\left \| I - I' \right \|_{2}^{2}}} \,\,\, .
\end{aligned}
\end{equation}

The final objective function is then defined by:
\begin{equation}
\begin{aligned}
E(\theta_{\textit{P}}) = \lambda_{heatmap}E_{heatmap} & +\lambda_{PAF}E_{PAF} \\ 
& +\lambda_{pixel}E_{pixel}  \,\,\, ,
\end{aligned}
\end{equation}
where the first two terms force PolishNet to synthesize images containing the correct pose, and the last term helps PolishNet to maintain the original pressure image's shape information.

\newcolumntype{C}{>{\hsize=\dimexpr2\hsize+12\tabcolsep+\arrayrulewidth\centering\relax}X}
\newcolumntype{Q}{>{\hsize=\dimexpr1\hsize+7.61\tabcolsep+\arrayrulewidth\centering\relax}X}
\newcommand{\mc }[1] {\multicolumn{1}{Q}{\footnotesize#1}}
\newcommand{\mcs}[1] {\multicolumn{1}{C}{\small#1}}
\begin{table*}[!t]
\begin{center}
\caption{The area under the PCK curves and their standard deviations are presented for our proposed method and OpenPose only.} \begin{tabularx}{\textwidth}{CQQQQQQQ}
\Xhline{2\arrayrulewidth}
\mcs{}                & \mc{\textbf{Head \& Neck}}   & \mc{\textbf{R Shoulder}}  & \mc{\textbf{R Elbow}}  & \mc{\textbf{R Wrist}} & \mc{\textbf{R Hip}}   & \mc{\textbf{R Knee}}  & \mc{\textbf{R Ankle}}          \\  \hline
\mcs{\textbf{OpenPose only}}   & \mc{19.4 $\pm$ 27.1} & \mc{73.0 $\pm$ 19.1}      & \mc{55.6 $\pm$ 15.4}   & \mc{28.1 $\pm$ 14.4}  & \mc{67.2 $\pm$ 25.5}  & \mc{56.4 $\pm$ 30.9}  & \mc{37.8 $\pm$ 20.3}   \\  
\mcs{\textbf{Proposed Method}} & \mc{92.9 $\pm$ 19.2} & \mc{92.1 $\pm$ 19.7}      & \mc{89.7 $\pm$ 20.0}   & \mc{83.1 $\pm$ 24.4}  & \mc{93.9 $\pm$ 20.1}  & \mc{94.1 $\pm$ 18.4}  & \mc{93.6 $\pm$ 19.5}  \\ \Xhline{2\arrayrulewidth}
\mcs{}                & \mc{\textbf{Neck}}   & \mc{\textbf{L Shoulder}}  & \mc{\textbf{L Elbow}}  & \mc{\textbf{L Wrist}} & \mc{\textbf{L Hip}}   & \mc{\textbf{L Knee}}  & \mc{\textbf{L Ankle}}          \\  \hline 
\mcs{\textbf{OpenPose only}}   & \mc{79.2 $\pm$ 18.8} & \mc{73.6 $\pm$ 21.9}      & \mc{57.4 $\pm$ 20.1}   & \mc{30.1 $\pm$ 17.7}  & \mc{4.5 $\pm$ 7.4}    & \mc{55.4 $\pm$ 30.9}  & \mc{39.6 $\pm$ 21.0}   \\ 
\mcs{\textbf{Proposed Method}} & \mc{94.9 $\pm$ 18.1} & \mc{92.2 $\pm$ 19.7}      & \mc{91.3 $\pm$ 19.7}   & \mc{85.4 $\pm$ 23.4}  & \mc{93.6 $\pm$ 19.5}  & \mc{94.0 $\pm$ 18.9}  & \mc{93.2 $\pm$ 18.6}  \\ \Xhline{2\arrayrulewidth}
\end{tabularx}
\label{table:average_detection}
\end{center}
\end{table*}

\noindent\textbf{Implementation Details:}
We implement the pipeline using TensorFlow on an NVIDIA Titan XP GPU. The convolution kernels of the PolishNet module were $3 \times 3$ with a stride of $2$, and the LeakyReLu use a negative activation coefficient of $0.1$. Bigger kernel sizes interpolate the body shapes better at the cost of losing input-output image similarity. We use an Adam optimizer to train the pipeline for $50$ epochs with a batch size of $16$. We use a learning rate of $10^{-4}$, which we decay with a rate of $0.95$ for every $100$ update iterations. Finally, $\lambda_{PAF}$ and $\lambda_{heatmap}$ are both set to $1$, while $\lambda_{pixel}$ is empirically set to $0.2$ to allow PolishNet to focus more on reconstructing the vague pressure points of the body.

\section{Experiment Setup and Results}
\subsection{Data Preparation}
We used the PmatData dataset \cite{ostadabbas2014bed, PhysioNet} to train and test our pressure-based pose estimation approach. The pressure data have been recorded by the Force Sensitive Application (FSA) pressure mapping mattress. The mattress was equipped with $32\times64$ sensors, $1$ inch apart from each other. The recording was performed with a frequency of $1$ Hz for a pressure range of $0$-$100$ mmHg. $13$ subjects, with a height range of $169$-$186$ \textit{cm}, a weight range of $63$-$100$ \textit{Kg}, and an age range of $19$-$34$ years participated in the experiment of sleeping on the mattress in a total of $17$ unique pose.

The PmatData dataset does not contain the ground-truth join labels needed for training and evaluation purposes. Therefore, we developed and utilized a tool in MATLAB for annotating the body part locations, and subsequently labeled $1000$ pressure maps. We then implemented an annotation tool that automatically annotated the rest of the pressure maps for each subject and each posture using similarity in the image space based on the sum of squared errors. This was possible since each subject appears to be in a very similar (almost identical) posture with small variances in terms of general position during the majority of recording sessions given posture class.

Next, we applied a $3 \times 3 \times 3$ spatio-temporal median filter on the input data to remove the noise generated from occasional sensor values measuring unexpected values. We also removed $6$ frames recorded at the time of transition between sleeping poses, since in some cases, they did not show a clear image of the body.

\subsection{Performance Evaluation} \label{evaluation}
To evaluate the performance of our pipeline on the annotated data, we used the \textit{probability of correct keypoint} (PCK) evaluation metric, which is a measure of joint localization accuracy \cite{yang2013articulated}. Accordingly, we measure the distance between the detected and ground-truth keypoints, and if this distance is below a certain threshold, the keypoint in question is considered as true-positive. The threshold is defined as a fraction of the person's size, where the size is defined as the distance between the person's left shoulder and right hip \cite{andriluka20142d}. To perform a thorough evaluation of our method, we use a leave-one-subject-out cross-validation, where we leave $2$ subjects out for validation and use the remaining subjects for training.

\begin{figure}[!t]
\begin{center}
{\includegraphics[width=1\linewidth]{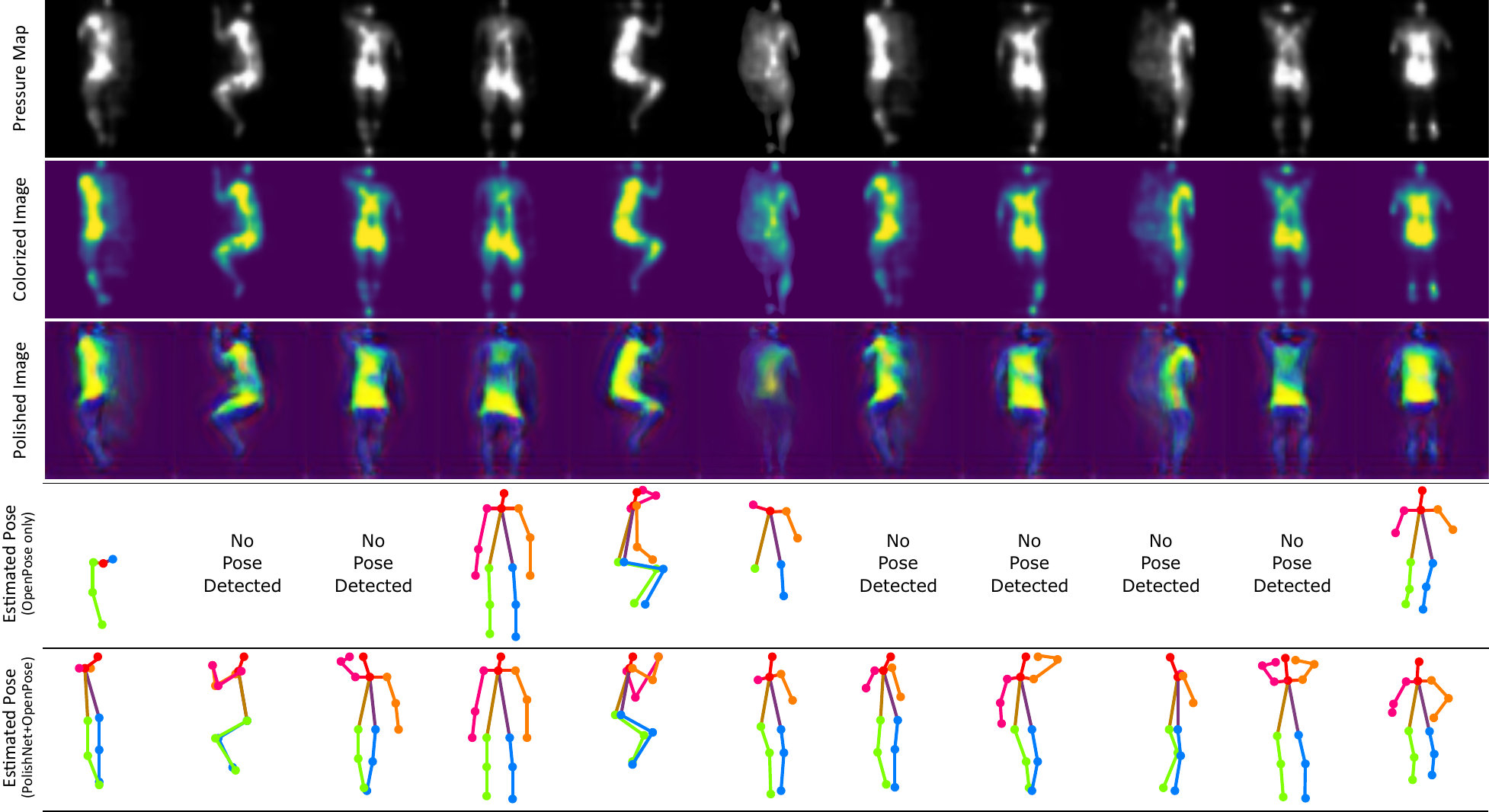}}
\end{center}
   \caption{Examples of the performance of the proposed architecture in estimating pose from input pressure maps.}
\label{fig:outputs}
\end{figure}

\begin{figure}[!t]
\begin{center}
\includegraphics[width=0.9\linewidth]{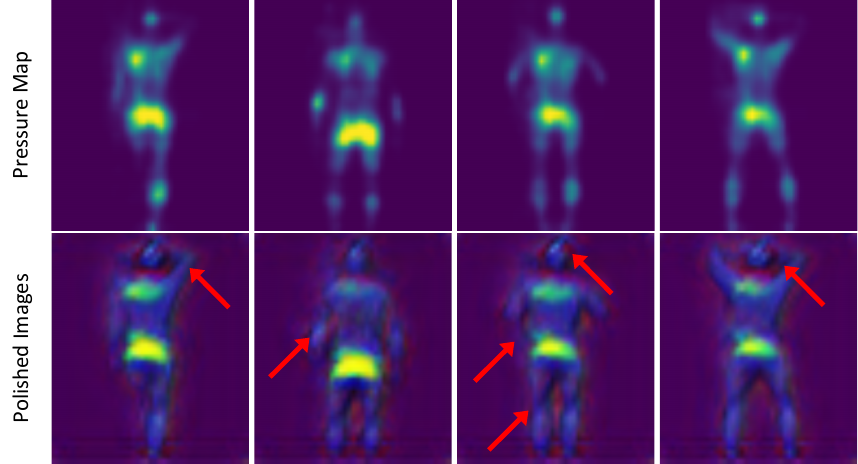}
\end{center}
   \caption{Illustration of examples where PolishNet has reconstructed weak body parts. Specifically note the arms, knees, and the head.}
\label{fig:long}
\label{fig:polished_images}
\end{figure}

We evaluate our method by comparing the performance of OpenPose on colorized pressure maps vs. the proposed pipeline, including PolishNet. The area under the PCK curves are presented in Table \ref{table:average_detection}, demonstrating that for all the body parts, our proposed pipeline considerably outperforms the use of only OpenPose on the colorized pressure maps. 
Low standard deviations in our test results indicate the consistency our model. It is observed from the table that for challenging body parts with weak pressure points, such as the wrists, or for ones with completely different appearances, such as the head, PolishNet provides consistently accurate images, eventually resulting in accurate pose estimation.

Some examples depicting the performance of our method are presented in Figure \ref{fig:outputs}. It is seen that in most cases, OpenPose alone is not able to correctly identify poses without PolishNet, if at all. Moreover, we observe that our proposed PolishNet + OpenPose pipeline accurately identifies the poses for vague input pressure maps. Since PolishNet is trained to synthesize images compatible with the image space by which OpenPose was trained, the polished outputs show a higher resemblance to common standing human poses. In Figure \ref{fig:polished_images}, we notice that PolishNet reconstructs and connects the limbs and weak pressure areas that are not clearly visible in the pressure maps. We have highlighted some of these reconstructed regions in Figure \ref{fig:polished_images}. Moreover, in some instances, PolishNet even attempts to interestingly synthesize \textit{outfits} for the subjects to make the output images look more natural and consistent with the input image space of the pose estimator. See Figure \ref{fig:outputs} row $3$, columns $3$, $4$, $8$, $10$, and $11$ for the synthesized outfit-like patches, especially around the hip and torso areas.

\newcolumntype{S}{>{\hsize=\dimexpr1\hsize+3.4\tabcolsep+\arrayrulewidth\centering\relax}X}
\newcommand{\ms}[1] {\multicolumn{1}{S}{\footnotesize#1}}
\begin{table}
\begin{center}
\caption{Quantitative evaluation of different models is presented. OpenPose and DeeperCut are tested with and without a PolishNet that has been \textit{pre-trained for OpenPose}. For both OpenPose and DeeperCut, the original and frozen versions are used.}
\begin{tabularx}{\columnwidth}{CSS}
\Xhline{2\arrayrulewidth}
\mcs{\textbf{Model}} & \ms{\textbf{Average Detection Rate}} \\ \Xhline{2\arrayrulewidth}
\mcs{OpenPose only}                     & \ms{47.7 $\pm$ 10.8} \\  
\mcs{PolishNet + OpenPose}              & \ms{\textbf{95.8 $\pm$ 0.3}} \\ 
\mcs{DeeperCut only}                    & \ms{54.1 $\pm$ 1.3} \\ 
\mcs{Pre-trained PolishNet + DeeperCut} & \ms{80.9 $\pm$ 2.4} \\  \Xhline{2\arrayrulewidth}
\end{tabularx}
\end{center} \label{table:summarized_results}
\end{table}

To further evaluate our method, we freeze PolishNet after training with OpenPose as the pose estimation module, then swap OpenPose with another popular pose estimation model, in this case, DeeperCut \cite{insafutdinov2016eccv, insafutdinov2017cvpr}. We then test the pipeline with pressure maps. The average area under the PCK curves for all the body parts and respective standard deviations are provided for each architecture in Table \ref{table:summarized_results}. As expected, PolishNet + OpenPose achieves the highest average detection rate since PolishNet is trained in a pipeline where OpenPose is used as the pose identification module. Interestingly, the pre-trained PolishNet followed by DeeperCut outperforms pose estimation with DeeperCut alone, improving the average detection rate by $26.8\%$. This further demonstrates that the images synthesized by PolishNet lie on, or close to, the manifolds with which most pose estimation models are trained. This allows us to use the pre-trained PolishNet as a learned pre-processor for enhancing ambiguous pressure maps, followed by any pre-trained pose identification block that may be selected based on available resources, constraints, and other properties.

\section{Conclusions}
Deep pose estimators are capable of detecting users' pose from natural images, while failing on data acquired from other devices such as pressure mapping systems, which are gaining popularity for health- and sleep-related research. In this paper, we addressed this issue by presenting a novel framework for in-bed pose estimation using an off-the-shelf pose estimation network, OpenPose, equipped with a learnable pre-processing block, called PolishNet. Using this design, our end-to-end model not only allows for pose estimation models to detect body parts with high accuracy, but also uses PolishNet to reconstruct vague and ambiguous body parts, such as wrists and knees. Furthermore, PolishNet results in synthesized images that can be used by other pose estimators as well. Our evaluation on a public dataset, PmatData, showed a $95.8\%$ detection rate with a leave-one-subject-out strategy, and $80.9\%$ when tested with another pose estimation network, namely DeeperCut. Finally, our proposed model can be effectively implemented for smart homes and clinical settings for ubiquitous and unobtrusive sleep monitoring.

\section{Acknowledgements}
The Titan XP GPU used for this research was donated by the NVIDIA Corporation. 

{\small
\bibliographystyle{IEEEbib}
\bibliography{main_bib}
}

\end{document}